\documentclass{article} % For LaTeX2e
\usepackage{iclr2015,times}
\usepackage{hyperref}
\usepackage{url}
\usepackage{amsmath}
\usepackage{amssymb}
\usepackage{graphicx}
\usepackage[tight,footnotesize]{subfigure}

\newcommand{\norm}[1]{\left\Vert#1\right\Vert}

\newcommand\vect[1]{{\bf#1}}
\newcommand\matr[1]{{\bf#1}}
\newcommand{\Real}{\mathbb R}
\newcommand\RR[1]{\mathbb{R}^{#1}}

\newtheorem{thm}{Theorem}

\newcommand{\rg}[1]{\textcolor{black}{#1}}

\title{On the Stability of Deep Networks }

\author{
Raja Giryes \& Guillermo Sapiro \\ %, and Alex M. Bronstein\\
Duke University \\ %and Tel Aviv University\\
\texttt{\{raja.giryes, guillermo.sapiro\}@duke.edu} \\ %;  bron@eng.tau.ac.il} \\
\And
Alex M. Bronstein \\
Tel Aviv University \\
\texttt{bron@eng.tau.ac.il} 
}

\iclrfinalcopy % Uncomment for camera-ready version

\begin{document}

\maketitle

\begin{abstract}

In this work we study the properties of deep neural networks (DNN) with random weights. We formally prove that these networks perform a distance-preserving embedding of the data. Based on this we then draw conclusions on the size of the training data and the networks' structure.
A longer version of this paper with more results and details can be found  in \citep{Giryes15Deep}. In particular,  we formally prove in \citep{Giryes15Deep} that DNN with random Gaussian weights perform a distance-preserving embedding of the data, with a special treatment for in-class and out-of-class data.
%A LONGER VERSION OF THIS PAPER ANALYZES ALSO THE ROLE OF THE TRAINING.

\end{abstract}

\section{Introduction}

Deep neural nets (DNN) have led to a revolution in the areas of machine learning, audio analysis,  and computer vision. Many state-of-the-art results have been achieved using these architectures. 
In this work we study the properties of these architectures with random weights. 
We prove that DNN preserve the distances in the data along their layers and  that this property allows stably recovering the original data from the features calculated by the network.
Our results provide insights into the outstanding empirically observed performance of DNN and the size of the training data. % and the role of weights learning.

Our motivation for studying networks with random weights is \rg{threefold}. 
First, one of the differences between the networks used two decades ago and state-of-the-art training strategies is the usage of random initialization of the weights. 
Second, a series of works \citep{Pinto09High,Saxe11random, Beyond11Cox}
empirically showed successful DNN learning techniques based on randomization.
\rg{Third, recent works that studied the optimization aspect in the training of deep networks also have done so via randomization \citep{Saxe14Exact, Dauphin14Identifying, Choromanska15Loss}.}

\citet{Bruna13Learning} show that the pooling stage in DNN causes a shift invariance property. \citet{Bruna14Signal} interpret this step as the removal of phase from a complex signal and show how the signal may be recovered after a pooling stage using phase retrieval methods.
In this short note, and for presentation purposes, we do not consider the previously studied pooling step, assuming the data to be properly aligned. We focus on the roles of the layers of a linear operation followed by an element-wise non-linear activation function.

\section{Stable Embedding of a Single Layer}
\label{sec:model}

We assume the input data to belong to a manifold $K$ with Gaussian mean width
\begin{eqnarray}
\omega(K) := E[ \sup_{\vect{x}, \vect{y} \in K} \langle \vect{g}, \vect{x} - \vect{y} \rangle ],
\end{eqnarray}
where the expectation is taken over $\vect{g}$ with normal i.i.d. elements. {In Section~\ref{sec:ent_net} we will illustrate this concept and exemplify the results with Gaussian mixture models (GMM).}

We say that $f : \Real \rightarrow \Real $ is a semi-truncated linear function if it is linear on some (possibly, semi-infinite) interval and constant outside of it, $f(0) = 0$, $0 < f(x) \le x, \forall x > 0$ and $0 \ge f(x) \ge x, \forall x < 0$. The popular rectified linear unit (ReLU), $f(x) = \max(0, x)$, is an example of such a function, while the sigmoid functions satisfy this property approximately. 
The following theorem shows that each standard DNN layer performs \rg{a stable embedding of the data in the Gromov-Hausdorff sense.}

\begin{thm}
\label{thm:deep_net_stable}
Let $\matr{M}$ be the linear operator applied at the $i$-th layer, $f$ the non-linear activation function, and $K \subset \mathbb{S}^{n-1}$ the manifold of the input data for the $i$-th layer. If $\sqrt{m}\matr{M} \in \RR{n \times m}$ is a random matrix with i.i.d normally distributed entries with $m=O(\omega(K)^2)$ being the output dimension, and
$f$ is a semi-truncated linear function, then with high probability 
\begin{eqnarray}
\norm{\vect{x} - \vect{y}}_2 \simeq d(f(\matr{M}\vect{x}), f(\matr{M}\vect{y} )), &  \forall \vect{x}, \vect{y} \in K,
\end{eqnarray}
\rg{where $d(\cdot, \cdot)$ is a variant of the Hamming distance that treats the positive values in the vectors as ones.
This result implies that the metric of the input data is preserved.}
\end{thm}

The proof follows from \citep{Plan14Dimension} and \cite{Klartag05Empirical}. % [Theorem~1.5] and [Theorem 1.4].

\cite{Mahendran14Understanding} demonstrate that it is possible to recover the input of DNN from their output. The next result provides a theoretical justification for their observation by  showing that it is possible to recover the input of each layer from its output:

\begin{thm}
\label{thm:deep_net_stable_rec}
Under the assumptions of Theorem~\ref{thm:deep_net_stable} there exists a program $\mathcal{A}$ such that 
\begin{eqnarray}
\norm{\vect{x} - \mathcal{A}(f(\matr{M}\vect{x} ) )}_2 \le \epsilon, 
\end{eqnarray}
where $\epsilon = O\left(\frac{\omega(K)}{\sqrt{m}}\right)$.
\end{thm}

The proof follows from \cite{Plan14Dimension}.

\section{Stable Embedding of the Entire Network}
\label{sec:ent_net}

In order to show that the entire network produces a stable embedding of its input, we need to show that the Gaussian mean width does not grow significantly as the data propagate through the layers of the network. 
Instead of bounding the variation of the Gaussian mean width throughout the network, we bound the change in the covering number $N(K,\epsilon)$, i.e., the lowest number of $\ell_2$-balls of radius $\epsilon$ that cover $K$.
Having the bound on the covering number, we use Dudley’s inequality \citep{Talagrand91Probability},
%\begin{eqnarray}
$\omega(K) \le C \int_0^\infty \sqrt{\log N(K, \epsilon)}d\epsilon$,
%\end{eqnarray}
to bound the Gaussian mean width variation, where $C$ is a constant. 

\begin{thm}
Under the assumptions of Theorem~\ref{thm:deep_net_stable},
%Let $\matr{M}$ be the linear operator applied at the $i$-th layer, $f$ the non-linear activation function, $K \in S^{n-1}$ be the manifold of the data in the input of the $i$-th layer and $\tilde{K}$ be the manifold of the data in the output. If ${\sqrt{m}}\matr{M} \in \RR{n \times m}$ is a random matrix with i.i.d. normally distributed entries and $f$ is a truncated linear function then
\begin{eqnarray}
N(f(\matr{M}K), \epsilon) \le N\left(K, \frac{\epsilon}{1+\frac{\omega(K)}{\sqrt{m}}}\right). 
\end{eqnarray}
\end{thm}

{\it Proof:} We now present a sketch of the proof, deferring the full proof that treats also the Gaussian mean width directly to a longer version of the paper. It is not hard to see that since a non-linear activation function shrinks the data, then it can not increase the size of the covering; therefore we focus on the linear part. 
Following \cite[Theorem 1.4]{Klartag05Empirical}, we have that the distances in $\matr{M}K$ are the same as the ones in $K$ up to a $1+\frac{\sqrt{m}}{\omega(K)}$ factor. This is sufficient to complete the proof. \hfill $\Box$

We demonstrate the implication of the above theorem for a GMM, i.e.,  $K$ consisting of $L$ Gaussians of dimension $k$ in the $\ell_2$-ball. For this model $N(K,\epsilon) = L\left(1 + \frac{2}{\epsilon} \right)^k$ for $\epsilon <1$ and $1$ otherwise (see \citet{Mendelson08Uniform}).
Therefore we have that $\omega(K) \le C'\sqrt{k + \log{L}}$  and that at each layer the Gaussian mean width grows at most with an order of $1+\frac{\sqrt{k} + \log{L}}{\sqrt{m}}$. Similar results can be shown for other models of union of subspaces and low dimensional manifolds.

\section{How Many Measurements Are Needed to Train the Network}
\label{sec:meas_num}

An important question in deep learning is what is the amount of labeled training samples needed at training. 
Using Sudakov minoration \citep{Talagrand91Probability}, one may get an upper bound on the size of an $\epsilon$-net in $K$. 
We have demonstrated that networks with random Gaussian weights realize a stable embedding; consequently, if a network is trained using the screening technique by selecting the best among many networks generated with random weights as suggested in \citet{Pinto09High,Saxe11random, Beyond11Cox}, then the number of data points needed to be used in order to guarantee that the network represents all the data is $O(\exp(\omega(K)^2 / \epsilon^2 ) )$. Since $\omega(K)^2$ is a proxy for the data dimension (see \cite{Plan14Dimension}), we conclude that the number of training points grows exponentially with the intrinsic dimension of the  data.

\section{Discussion and Conclusion}
\label{sec:conc}

We have shown that DNN with random Gaussian weights perform a distance-preserving embedding of the data. This result provides
%has several important implications on the design of DNN, and implies
 a relationship between the complexity of the input data and the size of the required training set. In addition, it draws a connection between the dimension of the features produced by the network,which still keep the metric information of the original manifold, and the complexity of the data. 
%This relation provides us with a better understanding of the structure of the network by the mean that from a certain stage the role of the layers in the network turns to be primarily calculating a transformation from the already learned feature space, with good metric,  to the relevant bit which should be turned on.

%In a longer version of this paper we will also present analysis of the effect of the training of the network. As random projections distort the distances in a uniform way, the training allows a distortion that shrinks intra-class distances and expels inter-class data points from each other. 
Though we have focused here on the case of DNN with linear filters with random Gaussian entries, it is possible to extend our analysis to distributions such as sub-Gaussian, and to random convolutional filters using proof techniques from \citep{Haupt10Toeplitz, Aperiodic12Saligrama, Rauhut12Restricted, Ai14One}. 
This and the extension to learned DNN will be presented in an extended version of this note.

%We showed that it is possible to view DNN as a stagewise metric learning process. Our results suggest that it might be possible to replace the current layers with other metric learning algorithms. 
%This stands in line with the recent literature on convolutional kernel methods (see \citep{Mairal2014Convolutional,Lu14How}).

%\begin{table}[t]
%\caption{Notations}
%\label{tbl:not}
%\begin{center}
%\begin{tabular}{ll}
%\multicolumn{1}{c}{\bf Symbol}  &\multicolumn{1}{c}{\bf Description}
%\\ \hline \\
%$\matr{X}_i \in \RR{d \times p_i}$         & $i$-th class \\
%$\vect{x}_{i,j} \in \RR{d}$             & $j$-th signals in the $i$-th class \\
%$N$             & Number of classes \\
%$p_i$             & Size of the $i$-th class \\
%$\matr{D}_i \in \RR{d \times n_i}$ & Dictionary representing class $i$ \\
%$k_i$ & Sparsity of the $i$-th class  \\
%$\alphabf_{i,j} \in \RR{n_i}$ & $k_i$-sparse representation of the signal $\vect{x}_{i,j}$\\
%\end{tabular}
%\end{center}
%\end{table}

\noindent 
{\bf Acknowledgments:}
This work is supported by NSF, DoD and ERC StG 335491.

\bibliography{../../deep_learn}
\bibliographystyle{iclr2015}

\end{document}